\documentclass{article}

\usepackage[T1]{fontenc}
\usepackage{lmodern}
\usepackage{cmap}

\usepackage{spconf}
\usepackage{amsmath}
\usepackage{amssymb}
\usepackage{graphicx}

\usepackage{enumitem}
\usepackage{subcaption}
\usepackage[hidelinks]{hyperref}

\newcommand{\cmark}{\checkmark}
\newcommand{\xmark}{\texttimes}

\title{InscriptionOCR: A Dataset and Method for Understanding Inscriptions}
\name{Jaidev Sanjay Khalane\qquad Akbar Ali \qquad V.~N. Prabhakar\qquad Shanmuganathan Raman}  
\address{
Indian Institute of Technology Gandhinagar, India.
}
\begin{document}
%
\maketitle
\begin{abstract}
Ancient script image restoration is a fundamental problem in computer vision, as it directly affects the reliable analysis and interpretation of historical documents and inscriptions.
Ashokan Brahmi is an ancient script extensively used during the reign of Emperor Ashoka in the 3rd century BC, primarily for inscriptions in Prakrit. These inscriptions, including major and minor rock and pillar edicts, constitute a valuable yet largely unexplored source of data for computational analysis. The degraded nature of inscription imagery and the lack of standardized digital resources pose significant challenges for automated processing.
 We present an end-to-end AI-based framework for understanding ancient inscriptions that encompasses image enhancement, optical character recognition (OCR), transliteration, and neural machine translation (NMT). The proposed pipeline processes low-quality images captured directly from stone inscriptions, performs image restoration and Brahmi script character recognition, maps the recognized characters to the Roman script, and finally translates the resulting Prakrit text into English. We also introduce two new datasets: (i) InscriptionOCR Dataset: the largest publicly usable digital OCR dataset for Brahmi script to date, consisting of over 200,000 character images across about 600 classes, and (ii) a bilingual Prakrit-English parallel corpus comprising over 2,000 sentence pairs for NMT. We believe that the proposed framework and datasets will facilitate future research in ancient script analysis, low-resource OCR, and digital epigraphy.
\end{abstract}

\section{Introduction}

Ancient inscriptions and documents \cite{dwivedi2020ocr} constitute a vital source of historical, linguistic, and cultural knowledge. Among these, the Ashokan inscriptions dating back to the 3rd century BC represent one of the earliest and most extensive uses of writing in the Indian subcontinent \cite{zhang2024llmco4mr}. These inscriptions, engraved on rocks and pillars across a vast geographical region, were primarily written in the Brahmi script and served as official proclamations of the Mauryan empire. Despite their historical significance, large-scale digitization and computational analysis of these inscriptions remain limited.

\begin{figure}[!t]
\centering
\includegraphics[width=0.5\textwidth]{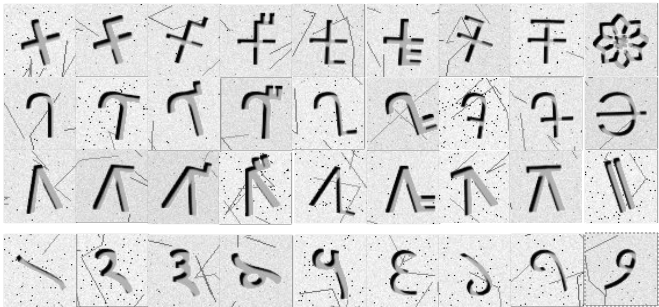}
\caption{
Representative dataset characters.
}

\label{fig:workflow}
\end{figure}

From a computer vision perspective \cite{he2016deep}, Ashokan inscriptions present several challenges. The available images are often captured under uncontrolled environmental conditions and are prone to erosion, surface damage, uneven illumination, shadows, biological growth, and background noise. Moreover, the Brahmi \cite{gunasekara2024deep} script exhibits significant variability in character shapes due to regional, temporal, and stylistic differences. The absence of standardized glyphs, combined with the scarcity of inscription-specific data, makes the direct application of existing optical character recognition \cite{mubarakkaa2024ocr}
(OCR) systems ineffective.

Recent advances in digital image processing and deep learning \cite{lecun2002gradient} have led to substantial progress in document analysis and OCR for modern scripts. However, ancient scripts such as Brahmi remain underexplored due to the lack of publicly available datasets that support reproducible research and fair benchmarking. Most existing studies focus on isolated components, such as preprocessing or character recognition, and do not provide end-to-end datasets that span from raw inscription images to machine-readable text representations.

In this work, we present a comprehensive dataset and processing pipeline for Ashokan Brahmi inscriptions. Starting from low-quality inscription images, we apply a series of classical image processing techniques, including grayscale conversion, denoising, thresholding, contour detection, and morphological operations, to enhance and restore the visual quality of the text. The restored images are then used for character-level segmentation and OCR, employing convolutional neural network-based character recognition models. Finally, the recognized Brahmi characters are mapped to their corresponding Roman transliterations, enabling standardized digital representation and facilitating downstream linguistic analysis \cite{zhang2024llmco4mr}.

The primary contribution of this paper is as follows
\begin{itemize}[topsep=2pt,itemsep=2pt,parsep=0pt,partopsep=0pt]
    \item A dataset-centric study bridging ancient epigraphical material with modern computer vision techniques.
    \item A comprehensive dataset named InscriptionOCR, specifically designed for facilitating OCR across inscription images, is also the largest Brahmi Characters dataset for OCR, both in terms of the number of classes and the number of images.
    \item One of the largest publicly usable datasets for the translation of text from the Prakrit language to the English language.
    \item An end-to-end framework to process inscription images in ancient script and language in order to return the overall content in a known language and script (English) for supporting future research in historical document analysis, and digital epigraphy.
\end{itemize}


\begin{table}[ht]
\centering
\renewcommand{\arraystretch}{1.2}
\setlength{\tabcolsep}{6pt}
\resizebox{0.48\textwidth}{!}{
\begin{tabular}{l c c c c}
\hline
\textbf{Author} &
\textbf{Translation} &
\textbf{Dataset Size} &
\textbf{OCR Classes} &
\textbf{Translation Dataset} \\
\hline
Gautam \cite{gautam2020dataset}        & \xmark & 7,011     & 170 & \xmark \\
Gunasekara \cite{gunasekara2024deep}   & \xmark & 250--300  & \xmark & \xmark \\
Mubarakkaa \cite{mubarakkaa2024ocr}   & \xmark & \xmark    & \xmark & \xmark \\
Dhivya \cite{dhivya2021tamizhi}       & \xmark & 190,000   & 209 & \xmark \\
Chaudhari \cite{chaudhari2024bridging}    & \cmark & \xmark   & \xmark & 1474 \\
\textbf{Our Work}  & \cmark & \textbf{253,800} & \textbf{564} & \textbf{2224} \\
\hline
\end{tabular}
}
\caption{Comparison of OCR and translation capabilities for Brahmi-based datasets.}
\label{tab:comparison}
\end{table}

\begin{table}[ht]
\centering
\renewcommand{\arraystretch}{1.25}
\setlength{\tabcolsep}{8pt}
\resizebox{0.48\textwidth}{!}{
\begin{tabular}{l c}
\hline
\multicolumn{2}{c}{\textbf{OCR Dataset}} \\
\hline
Total Images & 253800 \\
Training Images & 228,420 \\
Testing Images & 25,380 \\
\hline
\multicolumn{2}{c}{\textbf{OCR Class-wise Distribution}} \\
\hline
Numeric Characters & 30 \\
Vowel Characters & 17 \\
Punctuation Characters & 7 \\
Composite Characters (Plain + Voweled Consonants) & 510 \\
\hline
\multicolumn{2}{c}{\textbf{NMT (Translation) Dataset}} \\
\hline
Total Parallel Sentence Pairs & 2,224 \\
Mean Words per Sentence (Prakrit) & $\sim$10.6 \\
Mean Words per Sentence (English) & $\sim$20.8 \\
\hline
\end{tabular}
}
\caption{Specification of our  proposed dataset.}
\label{tab:our_dataset_spec}
\end{table}

\section{Related Work}

The OCR of the Brahmi script is still in its early stages, primarily due to the lack of large-scale annotated datasets and the script's complexity. Gautam \textit{et al.}~\cite{gautam2020dataset} began the work of creating a dataset of 7,011 isolated Brahmi characters in 170 classes, which served as the starting point for character-level recognition. Dhivya \textit{et al.}~\cite{dhivya2021tamizhi} designed an OCR system for historical Tamil Brahmi script using a large-scale dataset of approximately 190,000 images and experimented with various classification architectures.

Other authors have experimented with smaller, less well-documented datasets. Gunasekara \textit{et al.}~\cite{gunasekara2024deep} designed a manual OCR system using a few examples of characters, while Mubarakkaa \textit{et al.}~\cite{mubarakkaa2024ocr} worked on OCR and transliteration systems for Ancient Tamil Brahmi script without mentioning the size of the dataset used.

Apart from OCR, more recent work has focused on neural machine translation (NMT) for ancient Indian languages. Chaudhari \textit{et al.}~\cite{chaudhari2024bridging} developed an NMT system for the Maharashtri Prakrit language using multilingual transformer models, but these models require plain text inputs and do not handle the issues arising from the degraded images of inscriptions.

Other related works on ancient scripts of other civilizations include OCR for Akkadian and Sumerian cuneiform scripts~\cite{gordin2024cured} and end-to-end recognition and translation of Egyptian hieroglyphs~\cite{sobhy2023ai}, which show the increasing interest in deep learning-based analysis of ancient scripts. Table \ref{tab:comparison} compares the scope and extent of the related studies and our work in the Brahmi-Prakrit domain.

\begin{figure}
    \centering
    \includegraphics[width=0.85\linewidth]{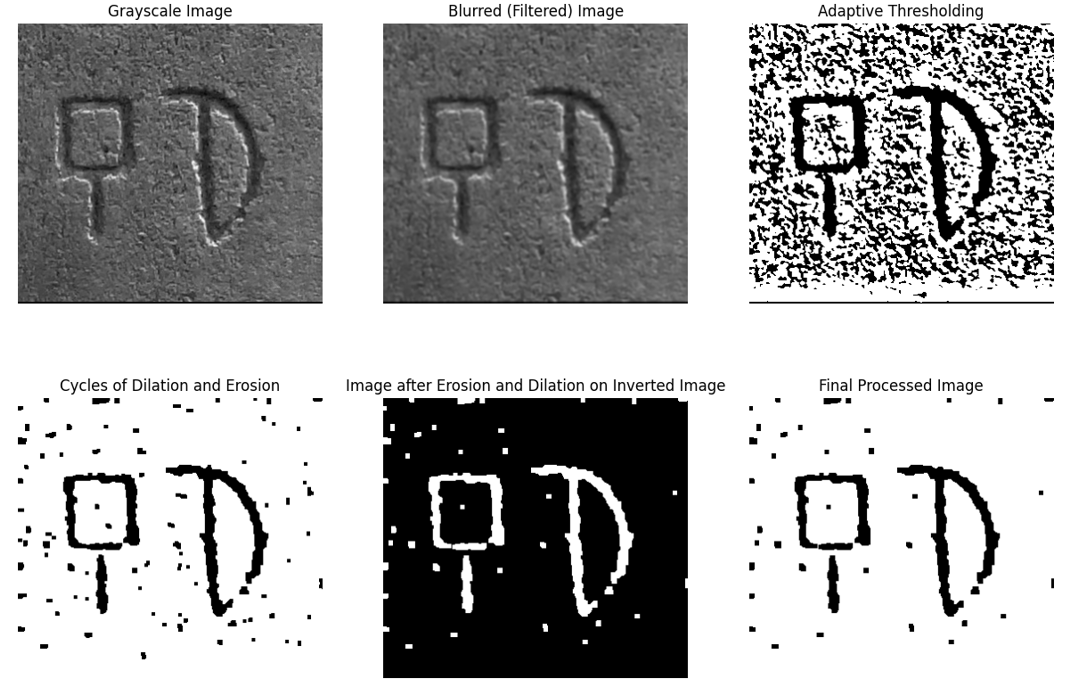}
    \caption{The Stepwise Results in Image Enhancement Phase.}
    \label{fig:f04gjhg}
\end{figure}

\section{Dataset}

\subsection{OCR Synthetic Dataset Generation}
To mitigate the lack of large-scale, annotated data for Brahmi script recognition, we propose an automated pipeline for generating a synthetic dataset that realistically simulates ancient stone inscriptions. 
The clean character images are first rendered using the Unicode-compliant font \textit{NotoSansBrahmi} on a $64 \times 64$-pixel canvas with center alignment. A stone-like background is then generated by sampling high-intensity pixel values with added Gaussian noise to simulate the granular texture of natural rock, followed by the addition of pitting artifacts and random cracks to simulate erosion and structural damage. The chiseled look of engraved characters is achieved through morphological stroke variations, beveling with spatial mask offsets, and gentle Gaussian blurring to simulate carving depth and long-term weathering. To further add variability, each sample is randomly rotated within $[-20^{\circ}, +20^{\circ}]$ and salt-and-pepper noise is added to simulate imperfectly carved inscriptions and imaging conditions.
The final images are composed by blending the chiseled character mask with the stone background, allowing the texture to remain visible within engraved regions. Using this process, we generate the InscriptionOCR dataset comprising 253{,}800 images (as shown in Figure~\ref{fig:workflow}) across 564 character classes, with 228{,}420 samples for training and 25{,}380 for testing.





\subsection{NMT Dataset}

The neural machine translation (NMT) dataset consists of 2,224 parallel sentence pairs between Prakrit and English, both represented in Roman script. The corpus was derived from the translations in \cite{hultzsch1991corpus}, which present multiple Ashokan inscriptions with section-wise parallel translations.

As the source material predates modern digital publishing standards, no reliable machine-readable version of the text was available. Consequently, the Prakrit and English texts were manually captured as images and processed using the Tesseract OCR engine for Roman script recognition. Since the inscriptions contain long passages, sentence-level parallel data were extracted using the original line-level delimiters employed by the author to indicate sentence alignment. These sentence pairs were manually verified and organized to form the initial parallel corpus.

To increase data diversity, the dataset was further augmented by introducing minor lexical variations through synonym substitution while preserving semantic consistency. The final dataset comprises 2,224 sentence pairs, of which 95\% were used for training and the remaining 5\% for testing. The precise specifications of the OCR and NMT datasets are displayed in Table \ref{tab:our_dataset_spec}. 

\section{Model Architectures and Metrics}

For the optical character recognition (OCR) task, several convolutional neural network (CNN) architectures were evaluated on the proposed Brahmi character dataset. The input images were grayscale with a resolution of $64 \times 64$, and the output layer contained 564 neurons corresponding to each character class. The LeNet-5 model \cite{lecun2002gradient} was employed as a lightweight baseline due to its simple architecture consisting of two convolutional layers followed by pooling operations and a sequence of fully connected layers. To explore deeper feature extraction, a VGG-16 inspired model \cite{simonyan2014very} was used, comprising thirteen convolutional layers arranged in five VGG-style blocks with roughly two to three convolutional layers each, followed by pooling, and three fully connected layers for classification. Other scalable architectures were also explored, including the ResNet-50 \cite{he2016deep} model based on the residual learning framework, as well as highly efficient models based on the EfficientNet \cite{tan2019efficientnet} architecture. Each of these OCR models was trained and tested on the synthetic Brahmi dataset, and detailed architectural configurations are provided in the accompanying source code.

For neural machine translation (NMT) from Prakrit (Roman script) to English, three pretrained transformer-based models were fine-tuned on the proposed parallel corpus. The Facebook M2M-100 (418M) model \cite{fan2021beyond}, capable of direct translation between 100 languages, including several Indian languages, was adapted via the Marathi-to-English variant for Prakrit translation. The T5-Base model \cite{raffel2020exploring}, a 220-million-parameter text-to-text transformer, was fine-tuned from the Hindi-to-English pretrained variant, while the Helsinki-NLP OPUS model \cite{tiedemann2023democratizing}, also pretrained for Hindi-to-English translation, was used due to the linguistic similarity between Hindi/Marathi and Prakrit. All NMT models were trained to generate English translations from the Romanized Prakrit sentences in the parallel corpus.

The performance of the OCR models was evaluated using the testing accuracy and a few other metrics, like precision, recall, and F1-score, computed on the held-out test set to estimate generalization performance. Translation quality for the NMT models was assessed using standard metrics, namely BLEU \cite{papineni2002bleu} and METEOR \cite{banerjee2005meteor} scores. 

\section{Methodology}

\subsection{Image Enhancement}

To obtain a clean image suitable for downstream analysis, the original inscription image was first converted to grayscale. The grayscale image was then denoised using a Gaussian filter. After denoising, histogram-adaptive thresholding was applied to binarize the image. Morphological operations, specifically sequences of dilation and erosion applied both to the thresholded image and its inverted version, were then performed to further clean the image. Finally, the processed image was inverted to produce a clean, enhanced version suitable for OCR.

To facilitate user-friendly processing, an image preprocessing and annotation tool was developed. This application provides adjustable parameters for each stage of the enhancement pipeline, including kernel sizes and threshold values, allowing users to control the sequence and intensity of each operation. The enhanced results are visualized immediately within the app interface. When the carvings are not adequately captured or the enhancement is unsatisfactory, the user can manually annotate the inscriptions using the annotator tool. The annotator interface allows the user to upload images, select pen size, and create annotations directly on the images. Once annotations are completed, they can be downloaded for use in subsequent stages. The detailed stepwise visualization is shown in Figure~\ref{fig:f04gjhg}.

\begin{figure*}[!t]
\centering
\includegraphics[width=0.95\textwidth]{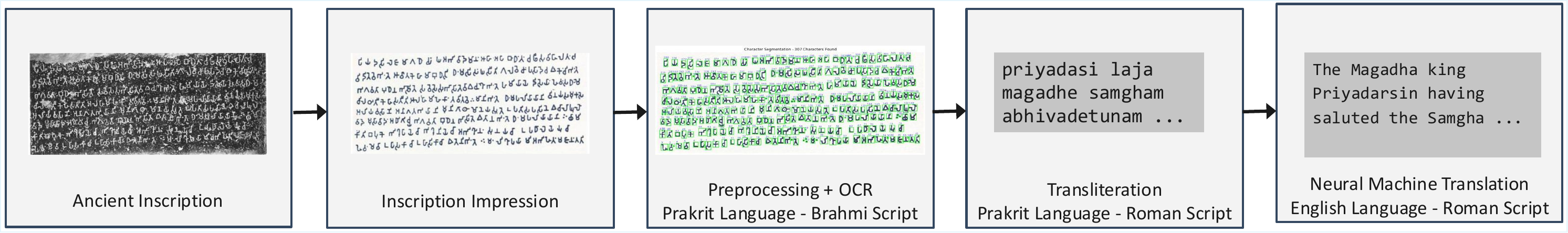}
\caption{
The figure illustrates the performance of the proposed method. Best viewed when \textbf{zoomed in}.
}
\label{fig:visualfsdlkfhsdksd}
\end{figure*}

\subsection{Preprocessing for Optical Character Recognition}

The annotated or enhanced images were further processed for OCR. Initially, all images were converted to grayscale (if not already), then binary thresholded, and morphologically cleaned using dilation and erosion. Contour analysis was then performed to identify regions corresponding to individual characters. Contours were filtered based on size and aspect ratio constraints to remove spurious detections, and the remaining contours were used to extract character blocks. These cropped character images were then resized to the input dimensions required by the CNN models and passed through the trained networks for classification. After classification, the detected characters were assembled into words and sentences by sorting them based on their coordinates and spacing. 

\subsection{CNN-Based Classification for OCR}

For character classification, four CNN models described in the previous section were trained on the Brahmi OCR dataset. Training was performed on a 16 GB Intel i5 (12th Gen) CPU using the parameters summarized in Table \ref{tab:cnn_params}. After training, the models were evaluated on the test dataset and subsequently used to infer the classes of individual characters extracted from inscription images.

\begin{table}[ht]
\centering
\renewcommand{\arraystretch}{1.2}
\setlength{\tabcolsep}{8pt}
\resizebox{0.48\textwidth}{!}{
\begin{tabular}{l c}
\hline
\textbf{Parameter} & \textbf{Value} \\
\hline
Batch Size & 32 \\
Optimizer & Adam (Adaptive Moment Estimation) \\
Learning Rate & 0.001 \\
Loss Function & Categorical Cross-Entropy \\
Number of Epochs & 20 \\
\hline
\end{tabular}
}
\caption{CNN training parameters.}
\label{tab:cnn_params}
\end{table}


\subsection{Transliteration}

Since the OCR outputs are in Brahmi script and Prakrit language, transliteration was required to match the Roman-script parallel dataset used for NMT. A direct mapping between Brahmi and Devanagari characters was used, followed by conversion to Roman script using the ISO transliteration scheme. After this step, the OCR outputs were represented in Romanized Prakrit, ready for translation.

\subsection{Neural Machine Translation}

For translating Romanized Prakrit to English, pretrained NMT models from Google (T5-Base), Facebook (M2M-100 418M), and Helsinki-NLP (OPUS) were fine-tuned on the parallel corpus described in the Dataset section. Fine-tuning was performed on a T4 GPU via Google Colab with the parameters summarized in Table \ref{tab:nmt_params}. After training, these models were used to generate English translations of the transliterated OCR outputs, completing the end-to-end integrated framework from raw image to translated text.
\begin{table}[ht]
\centering
\renewcommand{\arraystretch}{1.2}
\setlength{\tabcolsep}{8pt}
\resizebox{0.48\textwidth}{!}{
\begin{tabular}{l c}
\hline
\textbf{Parameter for NMT Model Finetuning} & \textbf{Value} \\
\hline
Learning Rate & $5 \times 10^{-5}$ \\
Batch Size & 8 \\
Weight Decay & 0.01 \\
Number of Training Epochs & 3 \\
\hline
\end{tabular}
}
\caption{NMT fine-tuning parameters.}
\label{tab:nmt_params}
\end{table}


\section{Results}

\begin{figure}
\centering
\includegraphics[width=3in]{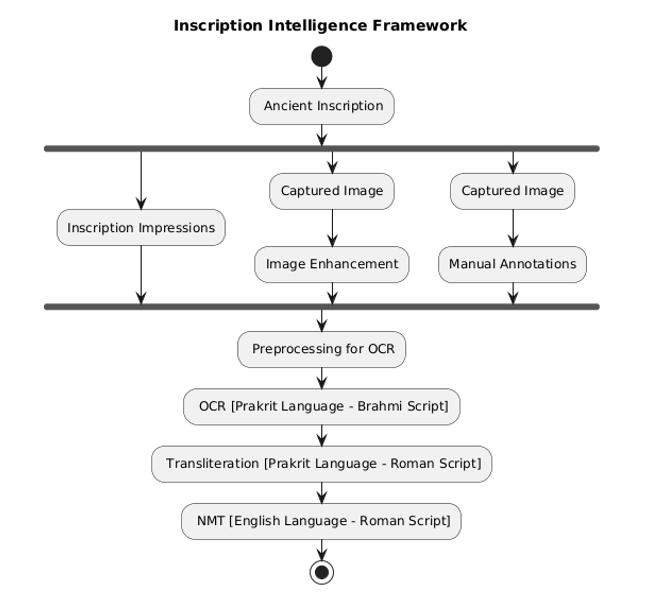}
\caption{Overall flow in our Inscription  Framework.}
\label{fig:f17}
\end{figure}

\subsection{End-to-End AI Integrated Framework}

In this study, we developed an end-to-end AI-based framework for analyzing and translating ancient inscriptions. The framework accepts input as either a captured image of an inscription or an impression of the inscription itself. The input is first processed through the image enhancement module or, if necessary, the manual annotation module to produce a clean binary image suitable for further analysis. The enhanced image is then preprocessed for OCR, including grayscale conversion, thresholding, morphological cleaning, and contour analysis to extract individual character regions. These regions are classified using CNN-based OCR models to extract Brahmi-script text in Prakrit. Subsequently, the OCR output is transliterated from Brahmi script to Roman script, making it compatible with the parallel dataset for Neural Machine Translation. Finally, the transliterated text is processed by fine-tuned NMT models to generate the corresponding English translation. The overall flow of this integrated framework is illustrated in Figure \ref{fig:f17}, and the stepwise visual outcomes are shown in Figure \ref{fig:visualfsdlkfhsdksd}.

\subsection{CNN-Based Classification for OCR}

The performance of the CNN-based OCR models was evaluated on the testing dataset, with results summarized in Table \ref{tab:cnn_model_comparison}. Among the four models, ResNet achieved the highest accuracy of 100.00\%, demonstrating superior performance in character classification. EfficientNet and VGG-16 also performed well, achieving test accuracies of more than 99\%, closely following ResNet. In contrast, LeNet achieved a slightly lower test accuracy of approximately 88\%, indicating that deeper or more optimized architectures are necessary for accurate recognition of complex Brahmi characters (Table~\ref{tab:cnn_model_comparison}).


\begin{table}[ht]
\centering
\renewcommand{\arraystretch}{1.2}
\setlength{\tabcolsep}{8pt}
\resizebox{0.48\textwidth}{!}{
\begin{tabular}{l c c c c}
\hline
\textbf{Model} & \textbf{Accuracy} & \textbf{Precision} & \textbf{Recall} & \textbf{F1 Score} \\
\hline
LeNet-5        & 0.8872 & 0.8912 & 0.8872 & 0.8753 \\
VGG-16         & 0.9939 & 0.9955 & 0.9939 & 0.9935 \\
EfficientNet   & 0.9996 & 0.9997 & 0.9996 & 0.9996 \\
ResNet-50      & \textbf{1.0000} & \textbf{1.0000} & \textbf{1.0000} & \textbf{1.0000} \\
\hline
\end{tabular}
}
\caption{Performance comparison of CNN-based OCR models.}
\label{tab:cnn_model_comparison}
\end{table}

\subsection{Neural Machine Translation}

The fine-tuned NMT models were evaluated on the test dataset using BLEU and METEOR scores, as shown in Table \ref{tab:nmt_results}. The Facebook M2M-100 418M model achieved the highest performance, with BLEU and METEOR scores of 56.367 and 0.772, respectively, significantly outperforming the other models. The Helsinki-NLP Opus model achieved a BLEU score of 14.120 and a METEOR score of 0.254, while the Google T5 Base model showed lower performance, with a BLEU score of 7.75 and a METEOR score of 0.093. These results indicate that the M2M-100 model is best suited for translating Romanized Prakrit into English within our framework (Table~\ref{tab:nmt_results}).
\begin{table}[ht]
\centering
\renewcommand{\arraystretch}{1.2}
\setlength{\tabcolsep}{8pt}
\resizebox{0.48\textwidth}{!}{
\begin{tabular}{l c c}
\hline
\textbf{Model Name} & \textbf{BLEU Score} & \textbf{METEOR Score} \\
\hline
M2M 100 418M & \textbf{56.367} & \textbf{0.772} \\
t5 Base & 7.750 & 0.093 \\
Helsinki-NLP Opus & 14.120 & 0.254 \\
\hline
\end{tabular}
}
\caption{NMT training results.}
\label{tab:nmt_results}
\end{table}


\subsection{ Conclusion}

In this study, we successfully achieved the dual objectives of creating extensive datasets for OCR of Brahmi characters and neural machine translation of Prakrit. Building on these datasets, we developed an end-to-end AI-based framework for translating ancient inscriptions, encompassing image enhancement, OCR, transliteration, and NMT. To facilitate practical use, we also implemented web applications that allow users to obtain translations directly from inscription images through a convenient interface. 

This study provides novel, large-scale datasets for Brahmi OCR (InscriptionOCR Dataset) and for Prakrit-to-English translation. Additionally, it presents a fully integrated AI framework designed for understanding ancient inscriptions. The framework and its applications provide valuable tools for scholars, linguists, and heritage preservationists. This represents a significant advancement towards the computational revival and analysis of ancient languages and scripts.

\bibliographystyle{IEEEbib}
\bibliography{strings,refs}

\onecolumn

\centering\textbf{Supplementary Material}
\addcontentsline{toc}{section}{Supplementary Material}
\label{sec:contributions}
\section{Dataset Documentation}

\begin{itemize}[label={}, leftmargin=30pt]
    \item \textbf{Description} \\
    This is the dataset for the Inscription Intelligence Project. The dataset contains two sections, one for the OCR of Brahmi Characters, also specialised towards inscriptions and the second one for NMT having parallel sentence pairs between Prakrit and English.

    \item \textbf{Code} \\
    \href{https://github.com/JaidevSK/InscriptionOCR-Brahmi-Prakrit}
    {GitHub repository}
    
    \item \textbf{Dataset} \\
    \href{https://www.kaggle.com/datasets/jaidevkhalane/inscriptionocr-brahmi-prakrit}
    {Kaggle dataset}
    \item \textbf{Data Subject(s)} \\ 
    Character Image Classification (OCR), Neural Machine Translation
    \item \textbf{Volume}\\ Approximately 650 MB.
    \item \textbf{Total Number of Images}\\ The OCR dataset comprises more than 250,000 images. The NMT dataset is in csv format with about 2000 sentence pairs.
    \item \textbf{Data Format}\\ OCR Data: png images, NMT Data: csv format.
   \item \textbf{Dataset Owner and Affiliation} \\
Indian Institute of Technology Gandhinagar, India.
    \item \textbf{Dataset Organization} \\ The OCR Data has separate folders for testing and training samples which again contain separate folders for every character, named by the character and contains the respective images. The NMT Data is in the form of a csv file with two columns, the first one being the Prakrit text in Roman Script and the second one being its corresponding translation in English.
    \item \textbf{License} \\ ATTRIBUTION-NONCOMMERCIAL 4.0 INTERNATIONAL
\end{itemize}

\section{Implementation Details}
 The supplementary document provides additional analyses in terms of various inscription datasets, restoration scenarios, and levels of degradation. Taken together, these analyses provide further insights into the performance characteristics of the proposed approach for restoring ancient inscription images.

\subsection{Dataset Generation}
For the generation of synthetic dataset similar to the that of ancient inscriptions, we use computer vision techniques to simulate how these ancient characters might appear when carved into stone and weathered by time.

The synthesis of the Brahmi dataset begins with the create\_character\_image function, which establishes a clean baseline by rendering characters onto a 64×64 pixel white canvas. To maintain linguistic accuracy, the NotoSansBrahmi font is utilized for proper Unicode support, while the usage of anchor ensures that the glyph remains centered for subsequent transformations. To move beyond digital perfection, the script simulates an authentic stone surface by generating a base texture with pixel values between 220 and 240, enriched with Gaussian noise, structural fissures via the generate\_cracks function, and porous "pitting" to mimic natural erosion.

The "chiseled" aesthetic is achieved through the apply\_stone\_augmentations function, where MinFilter and MaxFilter randomly erode or dilate strokes to replicate varying tool pressure. By offsetting the character mask to create beveling and applying a slight Gaussian blur to simulate centuries of weathering, the script transforms flat text into a recessed 3D engraving. To ensure the final machine learning model is robust, the pipeline
introduces dataset diversification through random rotations between
$[-20^\circ,+20^\circ]$, the addition of salt-and-pepper noise, and
complex composition techniques. This rigorous process ultimately produced a comprehensive dataset of more than 200,000 images as described previously across 564 classes.

\subsection{Image Enhancement}
\begin{figure}[htbp]
    \centering
    \begin{subfigure}[b]{0.37\linewidth}
        \centering
        \includegraphics[width=\linewidth]{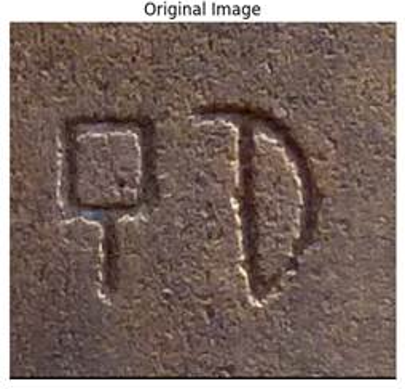}
        \caption{Original Image. Derived from: \cite{WikiImage_Budho}}
        \label{fig:f03}
    \end{subfigure}
    \hfill
    \begin{subfigure}[b]{0.56\linewidth}
        \centering
        \includegraphics[width=\linewidth]{Images/f04.png}
        \caption{Stepwise Results in Image Enhancement Phase}
        \label{fig:f04}
    \end{subfigure}
    \caption{Comparison of the Original Image and Image Enhancement Results}
    \label{fig:enhancement_comparison}
\end{figure}

In this subsection, we will display the figures corresponding to the image enhancement as described in the main text. Figure \ref{fig:f03} displays the original image, while Figure \ref{fig:f04} displays the results of intermediate steps such as Grayscaling, Filtering (Gaussian Blurring), Adaptive Thresholding, Cycles of Dilation, Erosion and Inversion as well as the Final Image. 

\label{sec:intro}

\subsection{Preprocessing for OCR}
\begin{figure}[htbp]
    \centering
    \begin{subfigure}[b]{0.3\linewidth}
        \centering
        \includegraphics[width=\linewidth]{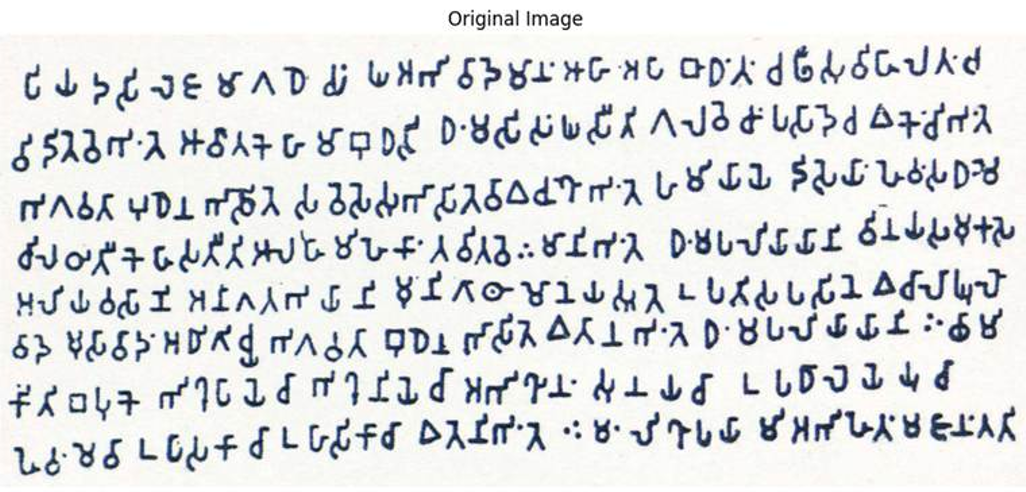}
        \caption{Original Image (Obtained as Inscription Impression) \cite{hultzsch1991corpus}}
        \label{fig:f13}
    \end{subfigure}
    \hfill
    \begin{subfigure}[b]{0.3\linewidth}
        \centering
        \includegraphics[width=\linewidth]{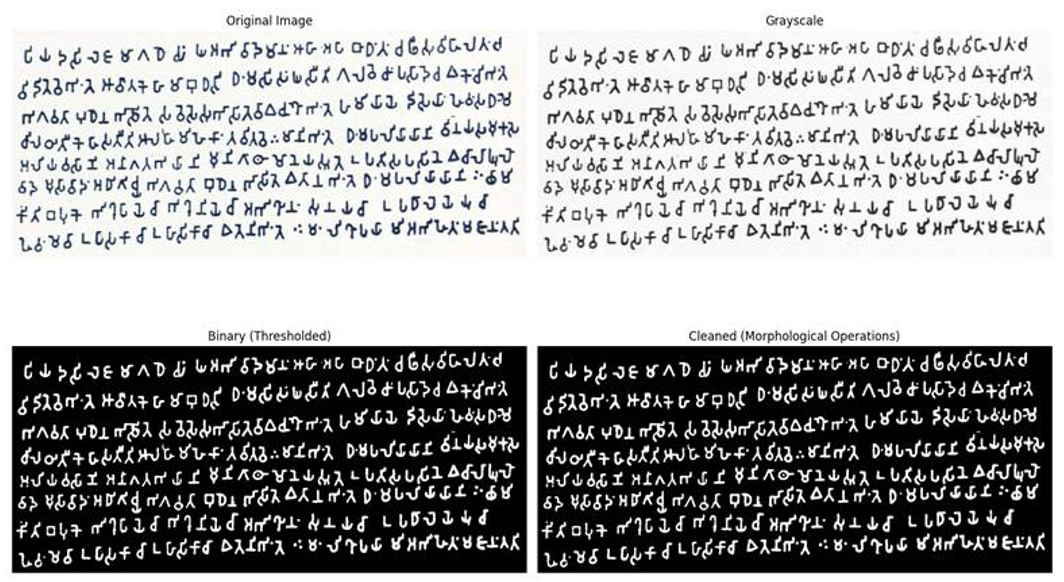}
        \caption{Preprocessing Steps before OCR}
        \label{fig:f14}
    \end{subfigure}
    \hfill
    \begin{subfigure}[b]{0.3\linewidth}
        \centering
        \includegraphics[width=\linewidth]{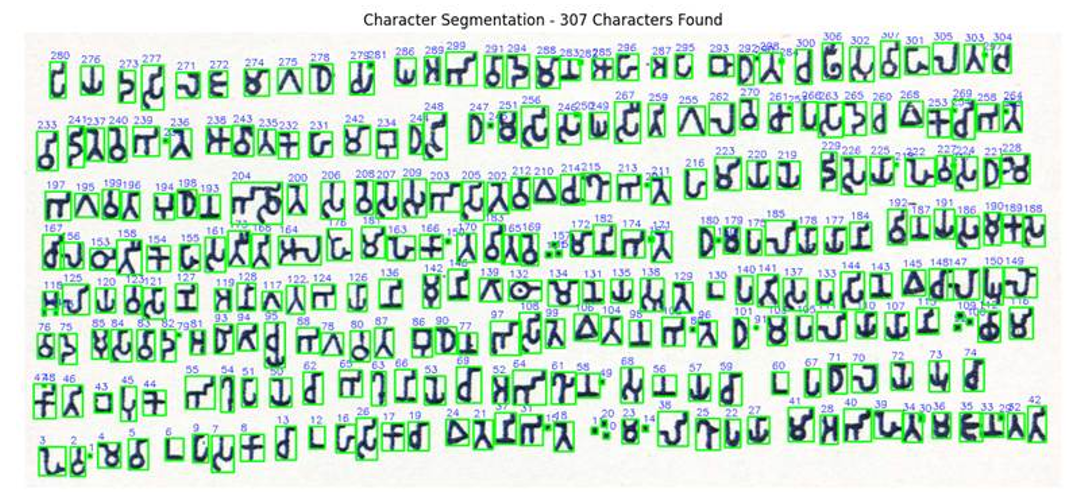}
        \caption{Contour Identification during OCR}
        \label{fig:f15}
    \end{subfigure}
    \caption{OCR Processing Pipeline for Inscription Images}
    \label{fig:ocr_pipeline}
\end{figure}

In this subsection, we present the intermediate results in the preprocessing required before the OCR step. Figure \ref{fig:f13} shows the initial image which was further prepared for OCR by performing Grayscaling, Thresholding and Cleaning through Morphological Operations like Erosion and Dilation as displayed in Figure \ref{fig:f14}. Figure \ref{fig:f15} displays the identified contours over the preprocessed image which will be cropped and passed to the CNN for further classification.

\section{Applications for Inscription Understanding}

\subsection{Image Enhancement}
\begin{figure}[htbp]
    \centering
    \begin{subfigure}[b]{0.43\linewidth}
        \centering
        \includegraphics[width=\linewidth]{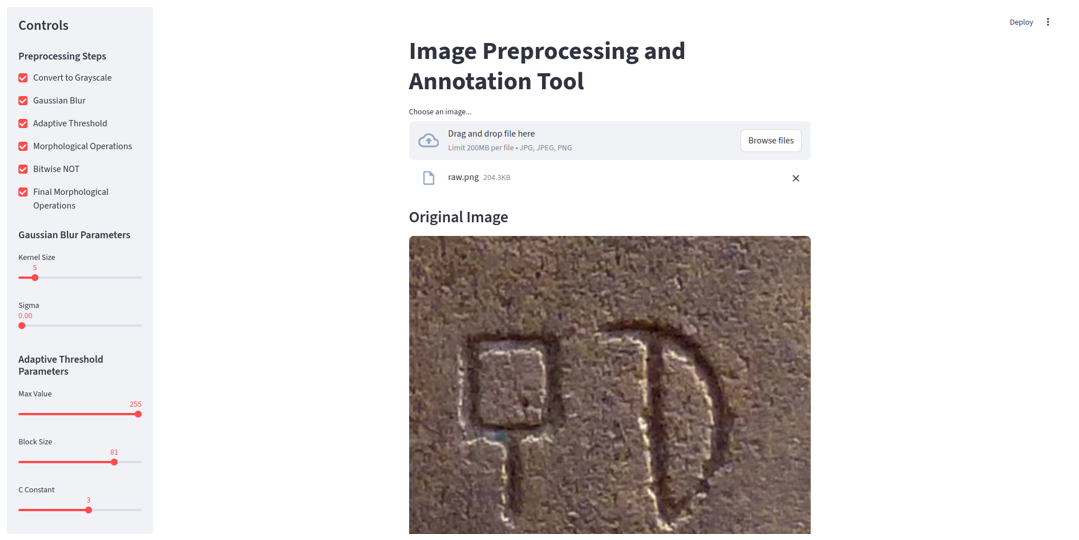}
        \caption{Image Enhancement Application Interface}
        \label{fig:f05}
    \end{subfigure}
    \hfill
    \begin{subfigure}[b]{0.43\linewidth}
        \centering
        \includegraphics[width=\linewidth]{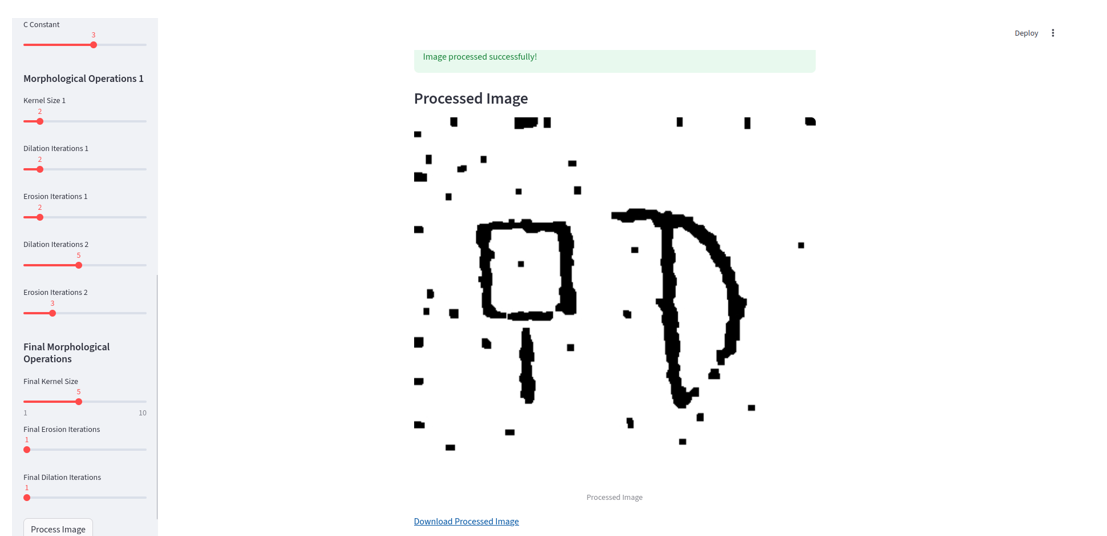}
        \caption{Image Enhancement Application Result}
        \label{fig:f06}
    \end{subfigure}
    \caption{Image Enhancement Application Interface and Output Result}
    \label{fig:app_interface_result}
\end{figure}

In order to make the image enhancement framework easy to use, a streamlit application was also developed as shown in Figure \ref{fig:f05}. The resultant image obtained can be downloaded by the user as shown in \ref{fig:f06}.

\subsection{Image Annotation Application}
\begin{figure}[htbp]
    \centering
    \begin{subfigure}[b]{0.55\linewidth}
        \centering
        \includegraphics[width=\linewidth]{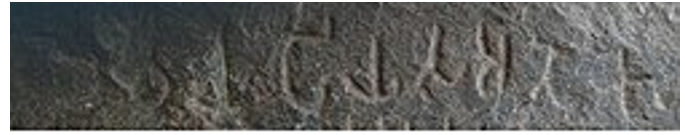}
        \caption{Original Image \cite{WikiImage_Asoka}}
        \label{fig:f07}
    \end{subfigure}
    \hfill
    \begin{subfigure}[b]{0.4\linewidth}
        \centering
        \includegraphics[width=\linewidth]{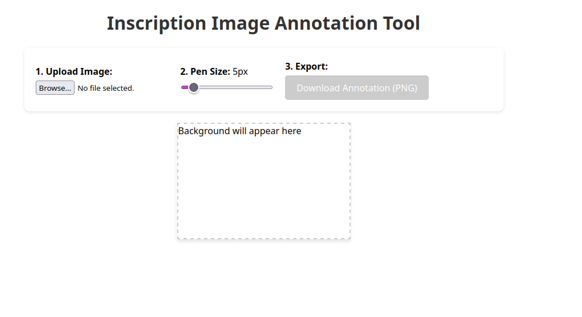}
        \caption{Annotation Application Interface}
        \label{fig:f08}
    \end{subfigure}

    \vspace{0.5em}

    \begin{subfigure}[b]{0.55\linewidth}
        \centering
        \includegraphics[width=\linewidth]{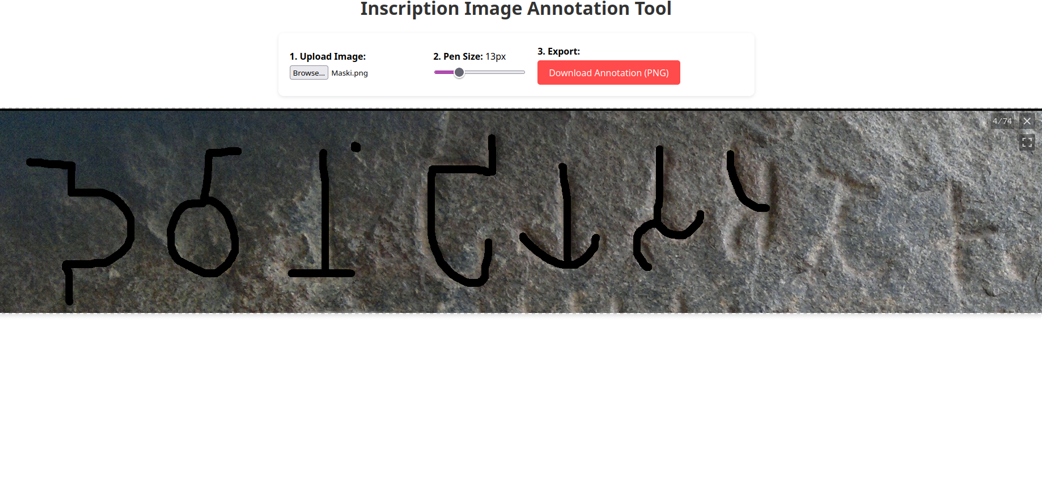}
        \caption{Image Being Annotated in the Application}
        \label{fig:f10}
    \end{subfigure}
    \hfill
    \begin{subfigure}[b]{0.4\linewidth}
        \centering
        \includegraphics[width=\linewidth]{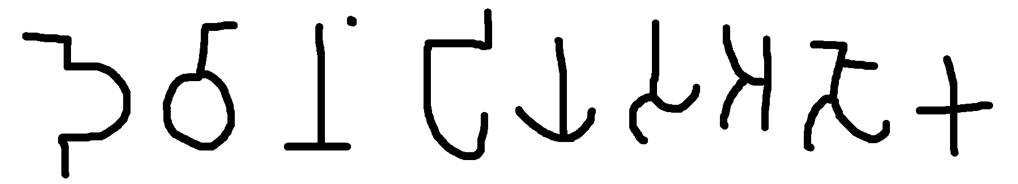}
        \caption{Downloaded Annotation}
        \label{fig:f12}
    \end{subfigure}

    \caption{Image Annotation Process and Results}
    \label{fig:annotation_process}
\end{figure}

In case the user is not satisfied with the image enhanced by the previous application, they can also manually annotate the image to trace the text written. Figure \ref{fig:f07} shows the original image. Figure \ref{fig:f08} shows the interface of our image annotation application. The image can then be uploaded to this application and the user can annotate this as shown in Figure \ref{fig:f10}. Finally, the user can download the annotation as in Figure \ref{fig:f12} and use it in the further steps.

\subsection{OCR+NMT Application}
\begin{figure}[htbp]
    \centering
    \begin{subfigure}[b]{0.55\linewidth}
        \centering
        \includegraphics[width=2in]{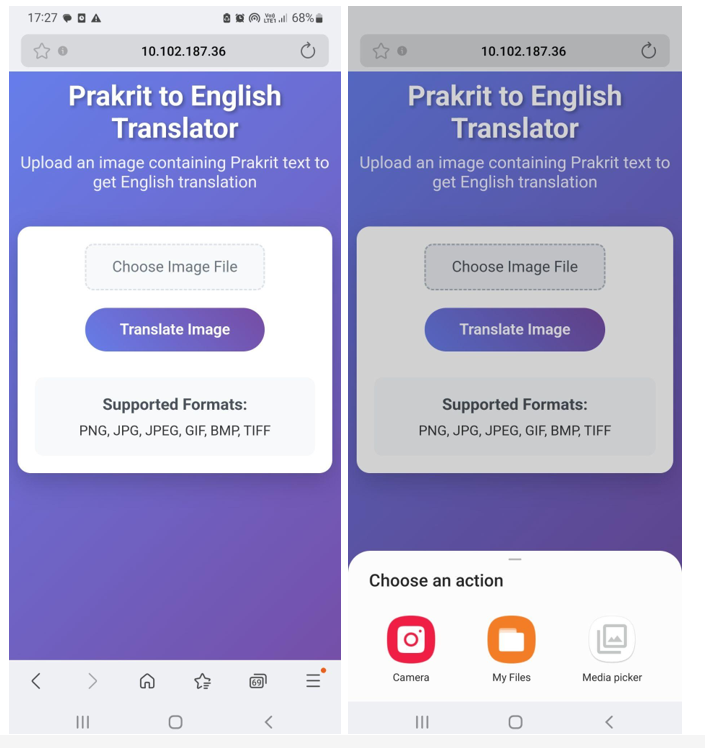}
        \caption{Mobile Application Interface for the Framework and Menu for Uploading the Image}
        \label{fig:f18}
    \end{subfigure}
    \hfill
    \begin{subfigure}[b]{0.4\linewidth}
        \centering
        \includegraphics[width=1in]{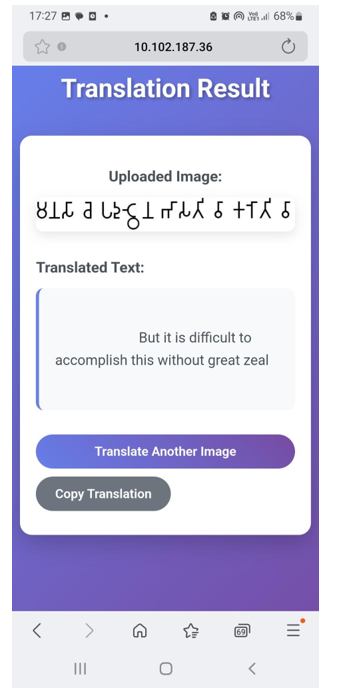}
        \caption{Overall Result of OCR + Transliteration + NMT (Mobile Screenshot)}
        \label{fig:f19}
    \end{subfigure}
    \caption{Mobile Application Interface and OCR+Transliteration+NMT Output}
    \label{fig:mobile_app_results}
\end{figure}

To make the integrated framework accessible to users, a web
application was developed using Flask (Python). This application allows users to upload enhanced images of inscriptions
and receive the corresponding English translation, integrating the OCR, transliteration, and NMT modules in a seamless
workflow. Since the computationally intensive operations are
performed on the server, users can access the application via
mobile devices, broadening its accessibility. The mobile interface and workflow for uploading and obtaining translations
are illustrated in Figs. \ref{fig:f18} and \ref{fig:f19}.

\end{document}